\documentclass[runningheads]{llncs}
\usepackage{cite}
\usepackage[utf8]{inputenc}
\usepackage[T1]{fontenc}
\usepackage{multirow}
\usepackage{mathrsfs}
\usepackage{graphicx}
\usepackage{color}
\usepackage{upgreek}
\usepackage{bbm}
\usepackage{url}
\usepackage{microtype}
\usepackage{amssymb}

\begin{document}
\title{DiffMix: Diffusion Model-based Data Synthesis for Nuclei Segmentation and Classification in Imbalanced Pathology Image Datasets}

\titlerunning{DiffMix for Nuclei Segmentation and Classification}

\author{Hyun-Jic Oh \and Won-Ki Jeong}
\authorrunning{H.J. Oh and W.K. Jeong}

\institute{Korea University, College of Informatics, \\
Department of Computer Science and Engineering \\
\email{wkjeong@korea.ac.kr}}
\maketitle
\begin{abstract}
\label{f1_abstract}
Nuclei segmentation and classification is a significant process in pathology image analysis. 
Deep learning-based approaches have greatly contributed to the higher accuracy of this task.
However, those approaches suffer from the imbalanced nuclei data composition, which shows lower classification performance on the rare nuclei class.
In this paper, we propose a realistic data synthesis method using a diffusion model.
We generate two types of virtual patches to enlarge the training data distribution, 
which is for balancing the nuclei class variance and for enlarging the chance to look at various nuclei.
After that, we use a semantic-label-conditioned diffusion model to generate realistic and high-quality image samples.
We demonstrate the efficacy of our method by experiment results on two imbalanced nuclei datasets, improving the state-of-the-art networks.
The experimental results suggest that the proposed method improves the classification performance of the rare type nuclei classification, 
while showing superior segmentation and classification performance in imbalanced pathology nuclei datasets.

\end{abstract}
\section{Introduction}
\label{f2_introduction}
In digital pathology, nuclei segmentation and classification are crucial tasks for the diagnosis of diseases. 
Due to its diverse nature (e.g., shape, size, and color) and large numbers, nuclei analysis in whole slide images (WSIs) is a challenging task where computerized processing has become a de facto standard these days~\cite{li2022comprehensive}.
With the advent of deep learning, many challenging problems in nuclei analysis, such as  color inconsistency, overlapping nuclei, and clustered nuclei, are effectively handled via data-driven approaches~\cite{raza2019micro, graham2019hover, zhao2020triple, doan2022sonnet}. 
Some of the recent work tackle nuclei segmentation and classification simultaneously. 
For example, HoVer-Net~\cite{graham2019hover} and SONNET~\cite{doan2022sonnet} perform nuclei segmentation using a distance map to identify nucleus instances, and then assign a proper class to each of them. 
Although such deep learning-based algorithms have shown promising performance and overcome various challenges in nuclei analysis, data imbalance among nuclei types in the training data has become a major performance bottleneck~\cite{doan2022gradmix}.

Data augmentation~\cite{simard2002transformation, chapelle2000vicinal} can be an effective solution to compensate for data imbalance and to generalize DNN by enlarging the learnable training distribution using virtual training data.
There exist several previous works for the image classification task. 
Mixup~\cite{zhang2017mixup} interpolates pairs of images and labels to generate virtual training data. 
CutOut~\cite{devries2017improved} randomly masks out square regions of input during training.
CutMix~\cite{Yun_2019_ICCV} cuts out patches from original images and pastes them onto other training images.
Recently, a generative adversarial network(GAN)~\cite{hou2019robust, gong2021style, lin2022insmix, wang2022sian} has been actively studied for pathology data augmentation.
However, training a GAN is a challenging procedure because of its instability and a need for hyper-parameter tuning~\cite{dhariwal2021diffusion}. 
Moreover, most of the previous works mainly focus on nuclei segmentation only without considering nuclei classification.
More recently, Doan~\textit{et al.}\cite{doan2022gradmix} proposed a data regularization scheme that addresses the data imbalance problem in pathology images.
The main idea is to cut the nuclei from a scarce class image and paste them onto the nuclei from an abundant class image. 
Since the source and target nuclei are different in size and shape, a distance-based blending scheme is proposed.
This method reduces the data imbalance problem to some extent, but it only considers pixel values for blending and some unrealistic blending artifacts can be observed, which is the main limitation of the method.

The main motivation for this work stems from the recent advances in generative models. 
Recently, the denoising diffusion probabilistic model(DDPM)~\cite{NEURIPS2020_4c5bcfec} has gained much attention due to its superior performance that surpasses that of conventional GANs %~\cite{muller2022diffusion} 
and has been successfully adopted to a conditional environment~\cite{nichol2021glide, dhariwal2021diffusion, wolleb2022diffusion}.
Among them, we were specifically inspired by Wang~\textit{et al.}~\cite{wang2022semantic}, 
the semantic diffusion model(SDM) which can synthesize a semantic image conditioned on the semantic label map. 
Since data augmentation for nuclei segmentation and classification requires accurate semantic image and label map pairs, we believe SDM fits well the data augmentation scenario of our imbalanced nuclei data while allowing much more realistic pathology image generation compared to the pixel-blending or GAN-based prior work.

\begin{figure}[t]
    \centering
    \includegraphics[width=0.99\textwidth]{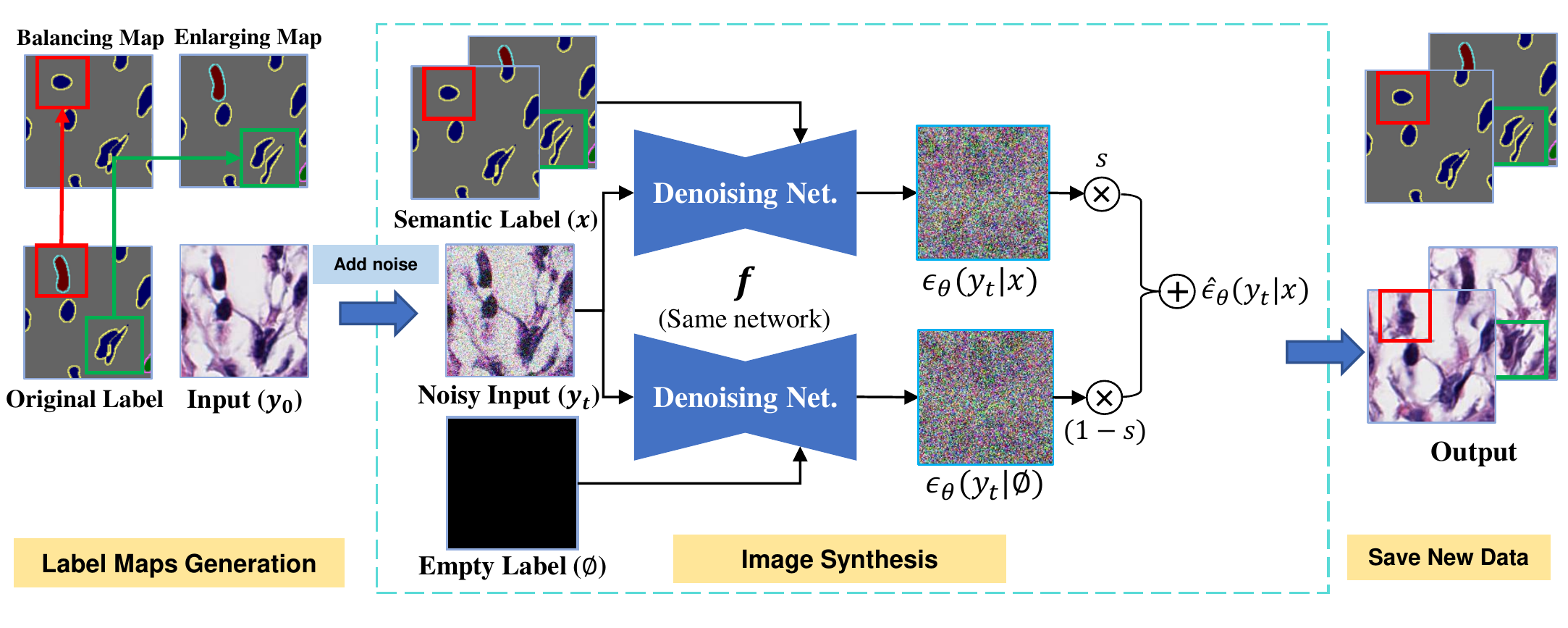}
    \caption{
    Framework of \texttt{DiffMix}. First, we generate custom semantic label maps($x$) and noisy images($y_t$). 
    Second, we synthesize image samples with a pretrained semantic diffusion model conditioned on the custom masks. Semantic label $x$ is custom mask to enlarge the data distribution. Lastly, we can utilize the synthesized image and label pairs on training DNN for nuclei segmentation and classification.}
    \label{fig:framework}
\end{figure}

In this paper, we propose a novel data augmentation technique using a conditioned diffusion model, \texttt{DiffMix}, for imbalanced pathology nuclei datasets. 
\texttt{DiffMix} consists of several steps as follows. 
First, we train SDM with semantic map guidance that consists of instance and class-type maps. 
Next, we build custom label maps by modifying the existing imbalance label maps. 
We change nuclei labels and randomly shift locations of the nuclei mask so that the number of each class label is balanced as well as data distribution is expanded.
Finally, we synthesize more diverse, semantically realistic,  and well-balanced new pathology nuclei images using SDM conditioned on our custom label maps. 
The main contributions of our work are summarized as follows.
{
\begin{itemize}
    \item We introduce a data augmentation framework for imbalanced pathology image datasets which can generate realistic samples using semantic diffusion model 
    conditioned on two custom label maps which can enlarge the data distribution.
    \item We demonstrate efficacy and generalization ability of our scheme with the experiment results on two imbalanced pathology nuclei datasets GLySAC\cite{doan2022sonnet} and CoNSeP\cite{graham2019hover}, improving the performance of the state-of-the-arts networks. 
     \item Our experiments demonstrate that the optimal approach for data augmentation depends on the level of imbalance, with balancing sample numbers and enlarging the training data distribution being critical factors to consider. % \wk{anything else?}
\end{itemize}
}

\section{Method}
\label{f3_method}
In this section, we describe the proposed method in detail. 
\texttt{DiffMix} works with several steps.
First, we train SDM first on training data. 
Balancing label maps have many rare class labels, and enlarging label maps are composed of some randomly shifted nuclei included.
Lastly, using pre-trained SDM and custom label maps, we synthesize realistic data to train on imbalanced datasets.
Before we dive into \texttt{DiffMix}, we start with a brief introduction to SDM. 
The overview of the proposed method is shown in Fig~\ref{fig:framework}. 

\subsection{Preliminaries}
SDM is a conditional denoising diffusion probabilistic model (CDPM) conditioned on semantic label maps. 
Based on CDPM, SDM follows two fundamental diffusion processes \textit{i.e.,} forward and reverse process.
The reverse process is a Markov chain with Gaussian transitions. 
When the added noise is large enough, the reverse process is approximated by a random variable $\textbf{y}_T\sim{\mathcal{N}(0,\textbf{I})}$, defined as follows:
\begin{eqnarray}
    p_{\theta}(\textbf{y}_{0:T}|\textbf{x})=p(\textbf{y}_T)\prod_{t=1}^{T} p_\theta(\textbf{y}_{t-1}|\textbf{y}_t,\textbf{x}) \\
    p_{\theta}(\textbf{y}_{t-1}|\textbf{y}_t, \textbf{x})=\mathcal{N} (\textbf{y}_{t-1}; \mu_\theta(\textbf{y}_t, \textbf{x}, t), 
    \Sigma_\theta(\textbf{y}_t, \textbf{x}, t))
\end{eqnarray}
The forward process implements Gaussian noise addition for $T$ timesteps based on variance schedule $\{$${\beta_1,...\beta_T}$$\}$ as below:
\begin{equation}
    q(\textbf{y}_t|\textbf{y}_{t-1})=\mathcal{N}(\textbf{y}_t;\sqrt{1-\beta_t}\textbf{y}_{t-1},{\beta_t}\textbf{I})
\end{equation}
With $\alpha_t:=1-\beta_t$ and $\bar{\alpha_t}:=\prod_{s=1}^{t}\alpha_s$, we can write the marginal distribution as follows,
\begin{equation}
    q({\textbf{y}_t}|\textbf{y}_0)=\mathcal{N}(\textbf{y}_t;\sqrt{\bar\alpha_t}\textbf{y}_0,(1-\bar\alpha_t)\textbf{I})
\end{equation}
The conditional DDPM is optimized to minimize the negative log-likelihood of the data for the given input and condition information.
If noise in the data follows Gaussian distribution with a diagonal covariance matrix $\Sigma_\theta(\textit{y}_t,\textit{x},t)=\sigma_t\textbf{I}$,
denoising can be the optimization target by removing the noise assumed to be present in data as follows,
\begin{equation}
    \mathcal{L}_{t-1}=\mathbb{E}_{y_0,\epsilon}[||\epsilon-\epsilon_\theta(\sqrt{\alpha_t}\textbf{y}_0+\sqrt{1-\alpha_t}\epsilon,\textbf{x},t)||_2]
\end{equation}

\subsection{Semantic Diffusion Model (SDM)}

SDM is a U-Net-based network that estimates the noise from the noisy input image.
Unlike other conditional DDPMs, the denoising network of SDM processes the semantic label map $x$ and noisy input $y_t$ independently. 
While $y_t$ is fed into the encoder part, $x$ is injected into the decoder to fully leverage the semantic information\cite{wang2022semantic}.
As for training, SDM is trained in a manner similar to the improved DDPM~\cite{nichol2021improved} so that it not only predicts the involved noise to reconstruct the input image but also predicts variances to enhance the log-likelihood of the generated images.
To improve sample quality, SDM utilized classifier-free guidance for inference.

SDM replaces the semantic label map $x$ with an empty (null) map $\emptyset$ in order to separate the noise estimated under the label map guidance by 
$\epsilon_\theta(y_t|x)$, from the noise estimated in an unconditioned case $\epsilon_\theta(y_t|\emptyset)$.
The strategy allows for the inference of the gradient of the log probability, expressed as follows,
\begin{equation}
    \epsilon_\theta(y_t|x)-\epsilon_\theta(y_t|\emptyset)\propto\nabla_{y_t}\log{p}(y_t|x)-\nabla_{y_t}\log{p}(y_t)\propto\nabla_{y_t}\log{p}(x|y_t)
\end{equation}
In the sampling process, the disentangled component $s$ is increased to improve the samples from conditional diffusion models, formulated as follows,
\begin{equation}
    \hat\epsilon(y_t|x)=\epsilon_\theta(y_t|x)+s\cdot(\epsilon_\theta(y_t|x)-\epsilon_\theta(y_t|\emptyset))
    \label{SDM_CFG}
\end{equation}
\subsection{Custom Semantic Label Maps Generation}

Fig~\ref{fig:framework} illustrates the process of creating custom label maps to condition the semantic diffusion model for synthesizing
desired data based on the original input image label $y_0$. 
We have prepared custom semantic label maps to condition SDM to direct it to synthesize the data for our imbalanced datasets.
So, we have considered making two types of semantic label maps %that are useful in
to improve imbalanced datasets. 
First, balancing maps to balance the number of nuclei among different nuclei types.
GradMix~\cite{doan2022gradmix} have increased the fewest type nuclei in datasets by cutting, pasting, and smoothing for both images and labels. 
On the other hand, we only used their mixed labels for our experiment.
Second, enlarging maps to stretch the data distribution available in training. 
We randomly moved nuclei positions on semantic maps to synthesize diverse image patches with SDM by conditioning with the unfamiliar semantic maps to lead the diffusion model to generate as many various patches as possible.
\subsection{Image Synthesis}
We synthesize the virtual data with pretrained SDM conditioned on the custom semantic label maps $x$.
Fig~\ref{fig:framework} depicted the data sampling process in SDM.
Before putting the original image $y_0$ into diffusion net $f$, we added noise to $y_0$.
The semantic label and noisy image $y_t$ will be simultaneously used.
To synthesize virtual images, we built two label maps and trained a semantic diffusion model. Before we start data synthesis, we add noise
on the input image $y_0$. After that, we input $y_t$ and $x$ to the pretrained denoising network $f$, $y_t$ for the encoder, and $x$ for the decoder.
As SDM generates samples, it uses the empty label $\emptyset$ to generate unconditioned output.
The image is sampled from an existing but noised patch, depending on the pre-defined time steps.
By doing so, we generate patches that are conditioned on the custom semantic label maps.
In this process, we added noise on the input image $y_0$, so we input the noised input $y_t$ into the pretrained denoising network $f$, 
and following \cite{wang2022semantic}, we input the custom semantic label maps to decoder parts. Then the semantic label maps 
will condition SDM to synthesize image data that satisfies the label maps. 
We have sampled the new data with DDIM\cite{song2020denoising} process, which diminishes the sampling steps but with high-quality data.

\section{Experiments}
\label{f4_experiments}
\begin{table}[htb!]
\centering
\scriptsize{
\caption{Quantitative results on GLySAC and CoNSeP. We implemented our scheme on two state-of-the-art networks, compared with GradMix. From Dice to PQ metrics state segmentation performance, and the metrics from Acc to $\textit{F}^s$ indicate classification performance.
The highest score in the same network and dataset is highlighted in \textbf{bold}. 
\textcolor{red}{Red} indicates the cases whose performance is at least 
3\% higher than the other methods.
}

\label{table:1}
\renewcommand{\arraystretch}{1.5}
\setlength{\tabcolsep}{3.5pt}
\begin{tabular}{rc|ccccc|cccccc}
\hline
\multirow{2}{*}{\textbf{Dataset}} & \multirow{2}{*}{\textbf{Method}} &
\multicolumn{5}{|c|}{\textbf{Segmentation}} & \multicolumn{5}{|c}{\textbf{Classification}} \\
\cline{3-12}
 \multicolumn{1}{c}{} & \multicolumn{1}{c|}{} &
 \textbf{Dice} & \textbf{AJI} & \textbf{DQ} & \textbf{SQ} & \textbf{PQ} & \textbf{Acc} & \textbf{$\textbf{F}^E$} & \textbf{$\textbf{F}^L$} & \textbf{$\textbf{F}^M$} & \textbf{$\textbf{F}^S$}\\ \hline\hline % & \textbf{Time (s)} \\ \hline\hline
 
 \multirow{10}{*}{GLySAC}
 & HoVer-Net & 0.839 & 0.670 & 0.807 & 0.787 & 0.637 & 0.713 & 0.565 & 0.556 & 0.315 & - \\
 & GradMix & 0.839 & 0.672 & 0.809 & 0.789 & 0.640 & 0.703 & 0.551 & 0.551 & 0.320 & - \\
 & DiffMix-E & 0.838 & 0.669 & 0.806 & 0.789 & 0.640 & \textbf{0.719} & 0.572 & \textbf{0.560} & 0.321 & - \\
 & DiffMix-B & \textbf{0.840} & \textbf{0.673} & \textbf{0.811} & \textbf{0.791} & \textbf{0.642} & 0.697 & 0.573 & 0.519 & 0.304 & - \\
 & DiffMix & 0.837 & 0.669 & 0.806 & 0.790 & 0.639 & 0.716 & \textbf{0.582} & 0.541 & \textbf{0.324} & - \\
 \cline{2-12}
 & SONNET & 0.835 & 0.660 & 0.789 & 0.792 & 0.627 & 0.679 & 0.511 & 0.511 & 0.305 & - \\
 & GradMix & 0.835 & 0.658 & 0.787 & 0.790 & 0.625 & 0.680 & 0.506 & 0.509 & 0.312 & - \\
 & DiffMix-E & 0.837 & 0.662 & \textbf{0.793} & \textbf{0.793} & \textbf{0.631} & \textbf{0.700} & 0.533 & \textbf{0.524} & \textbf{0.334} & - \\
 & DiffMix-B & \textbf{0.839} & 0.661 & 0.788 & 0.792 & 0.627 & 0.687 & 0.530 & 0.507 & 0.300 & - \\
 & DiffMix & 0.837 & \textbf{0.663} & \textbf{0.793} & 0.791 & 0.630 & 0.694 & \textbf{0.538} & 0.513 & 0.312 & - \\
\hline
 \multirow{10}{*}{CoNSeP}
 & HoVer-Net & 0.835 & 0.545 & 0.636 & 0.758 & 0.483 & 0.799 & 0.588 & 0.490 & 0.204 & 0.478 \\
 & GradMix & \textbf{0.836} & 0.562 & \textbf{0.658} & 0.765 & 0.504 & 0.802 & 0.598 & \textbf{0.519} & 0.144 & 0.494 \\
 & DiffMix-E & 0.832 & 0.550 & 0.645 & 0.760 & 0.492 & 0.804 & 0.602 & 0.486 & 0.223 & 0.493 \\ 
 & DiffMix-B & 0.835 & 0.558 & 0.653 & 0.762 & 0.499 & 0.809 & 0.595 & 0.496 & 0.324 & 0.498 \\ 
 & DiffMix & \textbf{0.836} & \textbf{0.563} & \textbf{0.658} & \textbf{0.766} & \textbf{0.505} & \textbf{0.818} & \textbf{0.604} & 0.501 & 
 \textcolor{red}{\textbf{0.363}} & \textbf{0.508} \\ 
 \cline{2-12}
 & SONNET & 0.841 & 0.564 & 0.646 & 0.766 & 0.496 & 0.863 & 0.610 & 0.618 & 0.367 & 0.560 \\
 & GradMix & 0.840 & 0.561 & 0.639 & 0.764 & 0.489 & 0.861 & 0.600 & \textbf{0.639} & 0.348 & 0.555 \\
 & DiffMix-E & 0.842 & 0.567 & 0.648 & \textbf{0.767} & 0.498 & 0.860 & 0.604 & 0.600 & 0.374 & 0.557 \\
 & DiffMix-B & 0.842 & 0.562 & 0.636 & 0.765 & 0.488 & 0.857 & 0.600 & 0.606 & 0.335 & 0.557 \\
 & DiffMix & \textbf{0.844} & \textbf{0.570} & \textbf{0.649} & 0.766 & \textbf{0.499} & \textbf{0.873} & {\textbf{0.622}} & 0.627 & \textcolor{red}{\textbf{0.463}} & \textbf{0.575} \\
\hline
\end{tabular}
}
\end{table}

\subsection{Datasets}
 In this study, we have used two imbalanced nuclei segmentation and classification datasets for our experiment.
First, GLySAC~\cite{doan2022sonnet} consists of 59 H\&E images of size 1000$\times$1000 pixels, and split into 34 train images and 25 test images.
The GLySAC has 30875 nuclei, and grouped into 3 nuclei types which are 12081 lymphocytes, 12287 epithelial and 6507 miscellaneous, respectively.
Second, CoNSeP~\cite{graham2019hover} consists of 41 H\&E images of size 1000$\times$1000 pixels, and divided into 27 train images and 14 test images. 
The CoNSeP has 24319 nuclei in total, and composed of four nuclei classes, which are 5537 epithelial nuclei, 3941 inflammatory, 5700 spindle, and 371 miscellaneous nuclei.

\subsection{Implementation Details}
We used one NVIDIA RTX A6000 to train SDM, we trained SDM for 10000epochs. 
For data synthesis, we implemented DDIM-based diffusion process from 1000 to 100,
and we added noise on the input image to SDM, setting $T$ as 55.
In our scheme, $\emptyset$ is defined as the all-zero vector as same as \cite{wang2022semantic} and set $s=1.5$ when sampling both datasets. 
We implemented experiments on two baseline networks SONNET\cite{doan2022sonnet} and HoVer-Net\cite{graham2019hover}.
SONNET is implemented with Tensorflow version 1.15\cite{abadi2016tensorflow} as software framework with two NVIDIA GeForce 2080 Ti GPUs.
HoVer-Net is trained with PyTorch 1.11.0 as software framework with one NVIDIA GeForce 3090 Ti GPU.
We implemented 4-fold cross validation for SONNET, and 5-fold cross validation for HoVer-Net.
For fair comparison, we trained each process with same iterations, changing only the epoch numbers depending on the training set size. 
\texttt{DiffMix} and GradMix used all the original patches and the same numbers of synthesized patches
In case of DiffMix-B, it has original data and balancing map based patches. Likewise, DiffMix-E training set consists of enlarging patches with original training set.
Therefore, We trained each baseline network for 100epochs, 75epochs for DiffMix-B and -E, and trained 50epochs for GradMix and \texttt{DiffMix}.

\begin{figure}[t]
    \centering
    \includegraphics[width=0.99\textwidth]{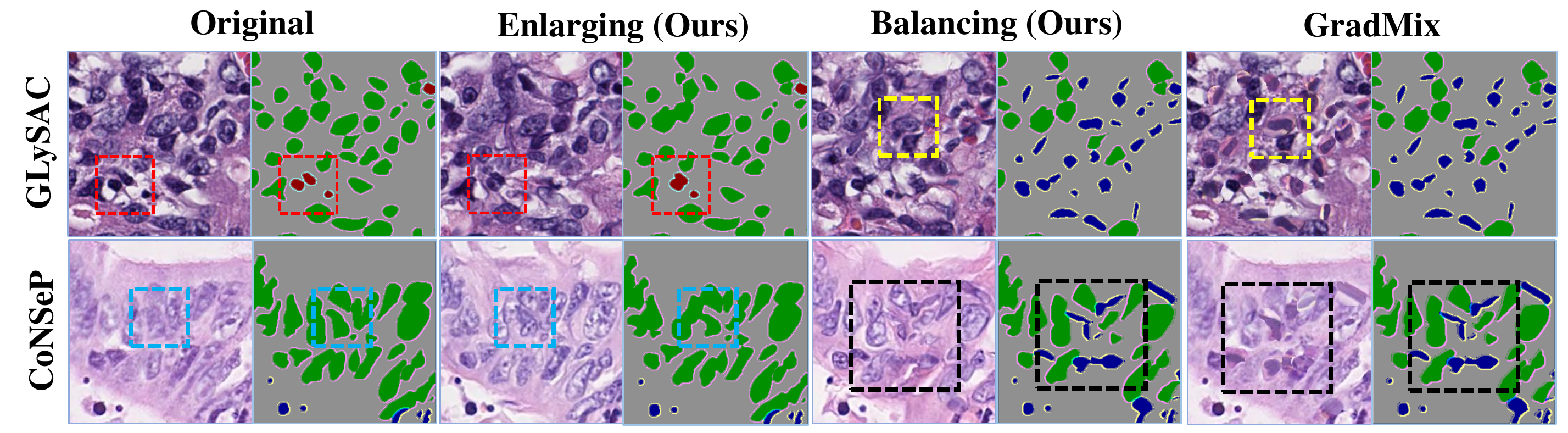}
    \caption{
    Qualitative comparison of synthesized patches. From left to right: Original data, enlarging map(ours), balancing map(ours), and GradMix, respectively. Top to bottom: GLySAC and CoNSeP datasets, respectively.
    Each image and its corresponding semantic label are paired together.
    In this work, we have utilized the same labels for balancing maps and GradMix. Comparing the result of balancing and GradMix patches, 
    our method generates semantically harmonized patches.
    }
    \label{table:patches}
\end{figure}

\subsection{Results}

Fig~\ref{table:patches} presents a qualitative comparison of synthesized patches. The original patch is on the left, followed by enlarging, balancing, and GradMix patches. Our two types of patches are well-harmonized with the surrounding structure, compared to GradMix. Moreover, using our scheme, we can synthesize many patches using a semantic diffusion model.

For the quantitative evaluation, we implemented two state-of-the-art networks, HoVer-Net and SONNET, on two public imbalanced nuclei-type datasets, GLySAC and CoNSeP.
Table~\ref{table:1} shows the results of 5 experiments per network for each dataset. We also conducted ablation studies on balancing (DiffMix-B) and enlarging (DiffMix-E) patch datasets. 
Before analyzing the experiment results, we computed the proportion of the least presenting nuclei type in each dataset. We found that Miscellaneous nuclei accounted for around 19$\%$ in GLySAC, but only 2.4$\%$ in CoNSeP. This means that GLySAC is more balanced in terms of nuclei types than CoNSeP.
Taking this information into account, we analyzed our experiment results. First, we noticed that DiffMix-E showed the highest classification performance in the GLySAC dataset. This result indicates that enlarging semantic map-based data synthesis, like DiffMix-E, had enough opportunities to enlarge its learning distribution for GLySAC. 
However, in the CoNSeP dataset, DiffMix-E performed lower than other methods, suggesting that if a dataset is somewhat balanced, it is important to enlarge the available data distribution.
\texttt{DiffMix} showed the highest performance in most metrics, with a 4$\%$ and 9$\%$ margin from the second-highest result in classifying Miscellaneous, successfully diminishing the classification performance variability among class types. Furthermore, \texttt{DiffMix} improved the segmentation and classification performance of two state-of-the-art networks, even compared to GradMix.

\section{Conclusion}
\label{f5_conclusion}
In this paper, we introduced \texttt{DiffMix}, a semantic diffusion model-based data augmentation framework for imbalanced pathology nuclei datasets. 
We have experimentally demonstrated that our method can synthesize virtual data which can balance and enlarge the imbalanced pathology nuclei datasets. 
Our method also outperforms the state-of-the-art GradMix in terms of qualitative and quantitative comparisons. 
Moreover, \texttt{DiffMix} enhances the segmentation and classification performance of two state-of-the-art networks, HoVer-Net and SONNET, even in imbalanced datasets like CoNSeP. 
Our results suggest that \texttt{DiffMix} can be used to improve the performance of medical image processing tasks in various applications. 
In the future, we plan to improve the performance of the diffusion model to generate various pathology tissue types. 

\bibliographystyle{splncs04}
\bibliography{f6_refs}

\clearpage
% Setting table caption for supplementary materials
\setcounter{table}{0}
\renewcommand{\thetable}{S\arabic{table}}
\setcounter{figure}{0}
\renewcommand{\thefigure}{S\arabic{figure}}

% supplementary
% contents: 
% data information(%), patch information(patch 개수), 학습 params, patch-T에 따른 생성되는 patch 차이, performance비교(overlay), SDM architecture

\begin{table}[htb!]
    \centering
    \scriptsize
    {
        \caption{This table shows the details of datasets. 
        In each dataset, Train and Test represent the information about the training set and test set. 
        Each cell shows the number and percentage (\%) of nuclei.}
        \renewcommand{\arraystretch}{1.5}
        \setlength{\tabcolsep}{3.3pt}
        \begin{tabular}{cc|ccccc} %c|cccccc}
        \hline
        \multicolumn{2}{l|}{\textbf{Dataset}} & \textbf{\textls[-125]{Lymphocyte/Inflammatory}} & \textbf{Epithelial} & 
        \textbf{Miscellaneous} & \textbf{Spindle} & \textbf{Total}
        \\ \hline\hline
        \multirow{3}{*}{GLySAC}
        & Train & 7409 (41.3\%) & 7154 (39.9\%) & 3386 (18.9\%) & - & 17949 \\
        & Test & 4672 (36.1\%) & 5133 (39.7\%) & 3121 (24.1\%) & - & 12926 \\
        & Total & 12081 (39.1\%) & 12287 (39.8\%) & 6507 (21.1\%) & - & 30875 \\
        \hline
        \multirow{3}{*}{CoNSeP}
        & Train & 3941 (25.3\%) & 5537 (35.6\%) & 371 (2.4\%) & 5700 (36.7\%) 
        & 15549 \\
        & Test & 1638 (18.7\%) & 3214 (35.6\%) & 561 (6.4\%) & 3357 (38.3\%) 
        & 8770 \\ 
        & Total & 5579 (22.9\%) & 8751 (36.0\%) & 932 (3.8\%) & 9057 (37.2\%) 
        & 24319 \\
        \hline
        \end{tabular}
    }
    \label{table:sup1_dataset_details}
\end{table}

\begin{figure}[hb!]
    \centering
    \includegraphics[width=0.99\textwidth]{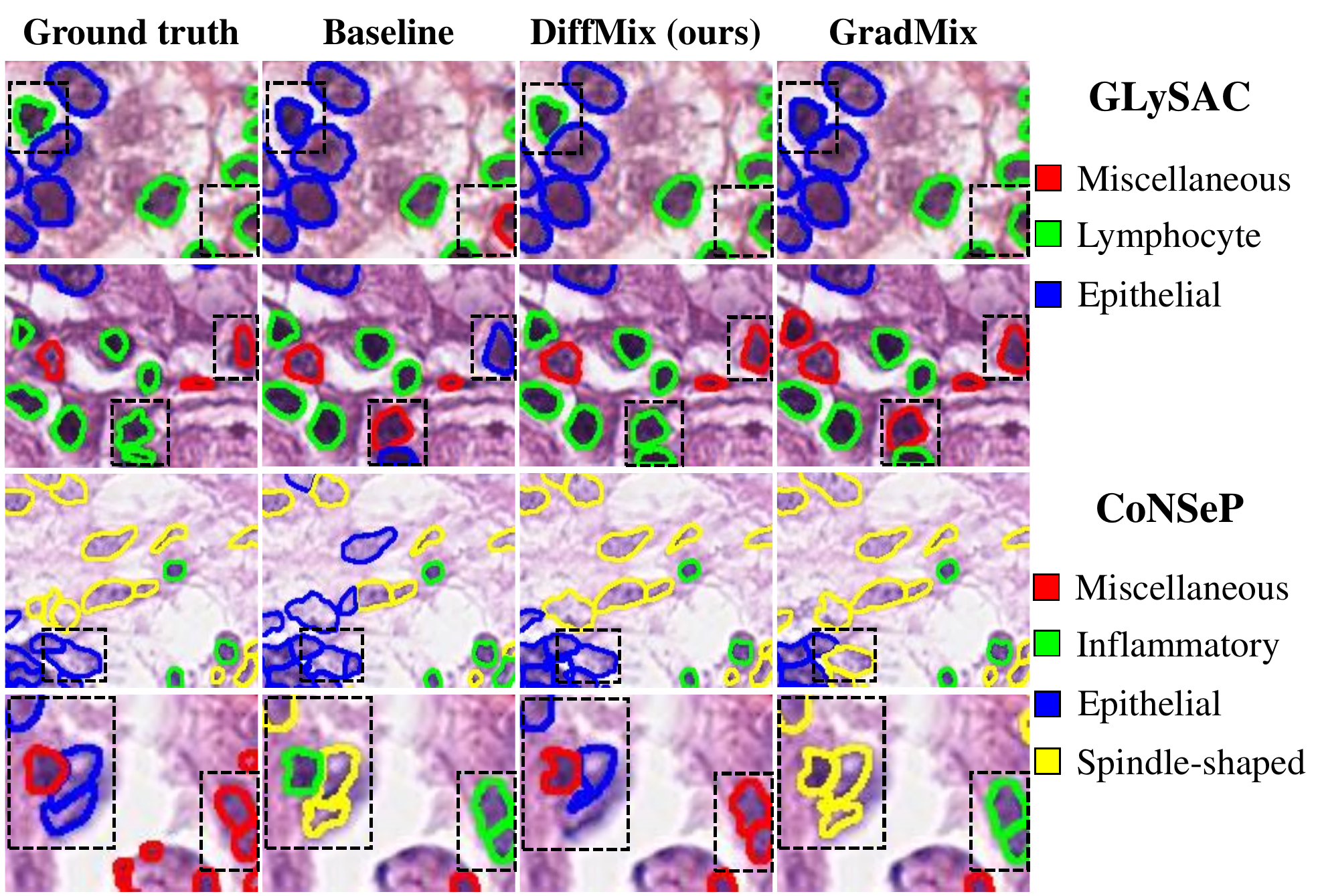}
    \caption{
        Qualitative results are presented from left to right: Ground truth, Baseline, DiffMix (ours), and GradMix, respectively. 
        The top two rows show the results of GLySAC, and the bottom two rows display the results of CoNSeP.
    }
    \label{table:qual_overlay}
\end{figure}

% \begin{table}[htb!]
%     \centering
%     \scriptsize{
%         \caption{
%             This table describes the details of metrics in Table~\ref{table:1}. 
%             Segmentation and Classification represent metrics for nuclei segmentation and classification, respectively.
%             }
%         \renewcommand{\arraystretch}{1.5}
%         \setlength{\tabcolsep}{4.8pt}
%         \begin{tabular}{cc|l}
%         \hline
%         \multicolumn{2}{l|}{\textbf{Metric}} & \textbf{Description} 
%         \\ \hline\hline
%         \multirow{5}{*}{\textbf{Segmentation}}
%         & \textbf{Dice} & Dice coefficient \\
%         & \textbf{AJI} & Aggregated Jaccard Index (AJI) \\
%         & \textbf{DQ} & Detection Quality \\
%         & \textbf{SQ} & Segmentation Quality \\
%         & \textbf{PQ} & Panoptic Quality \\
%         \hline
%         %
%         \multirow{5}{*}{\textbf{Classification}}
%         & \textbf{Acc} & Accuracy for each nuclei type \\
%         & \textbf{$\textbf{F}^E$} & F1-score for epithelial nuclei \\
%         & \textbf{$\textbf{F}^L$} & F1-score for lymphocyte\&Inflammatory nuclei \\
%         & \textbf{$\textbf{F}^M$} & F1-score for Miscellaneous nuclei \\
%         & \textbf{$\textbf{F}^S$} & F1-score for Spindle nuclei \\
%         \hline
        
%         % \multirow{5}{*}{\textbf{Seg.}}

%         \end{tabular}    
%     }
%     \label{}
% \end{table}

\begin{table}[htb!]
    \centering
    \scriptsize{
        \caption{
            This table describes the details of metrics in Table~\ref{table:1}. 
            \textbf{Seg.} and \textbf{Cla.} represent metrics for nuclei segmentation and classification, respectively.
            }
        \renewcommand{\arraystretch}{1.8}
        % \setlength{\tabcolsep}{4.8pt}
        % \specialrule{0.05em}{\aboverulesep}{0pt}
        % \specialrule{0.05em}{\belowrulesep}{0pt}
        \begin{tabular}{ll|ll}
        \hline
        \textbf{Seg.} & \textbf{Description} & \textbf{Cla.} & \textbf{Description}
        \\ 
        \hline\hline
        \textbf{Dice} & Dice coefficient & \textbf{Acc} & Accuracy for each nuclei type \\
        \textbf{AJI} & Aggregated Jaccard Index (AJI) & \textbf{$\textbf{F}^E$} & F1-score for epithelial nuclei \\
        \textbf{DQ} & Detection Quality & \textbf{$\textbf{F}^L$} & F1-score for lymphocyte/Inflammatory nuclei \\
        \textbf{SQ} & Segmentation Quality & \textbf{$\textbf{F}^M$} & F1-score for Miscellaneous nuclei \\
        \textbf{PQ} & Panoptic Quality & \textbf{$\textbf{F}^S$} & F1-score for Spindle-shaped nuclei \\
        \hline
        \end{tabular}
    }
    \label{}
\end{table}

%% figure 2 - patch (zoom-in) images
\begin{figure}[hb!]
    \centering
    \includegraphics[width=0.99\textwidth]{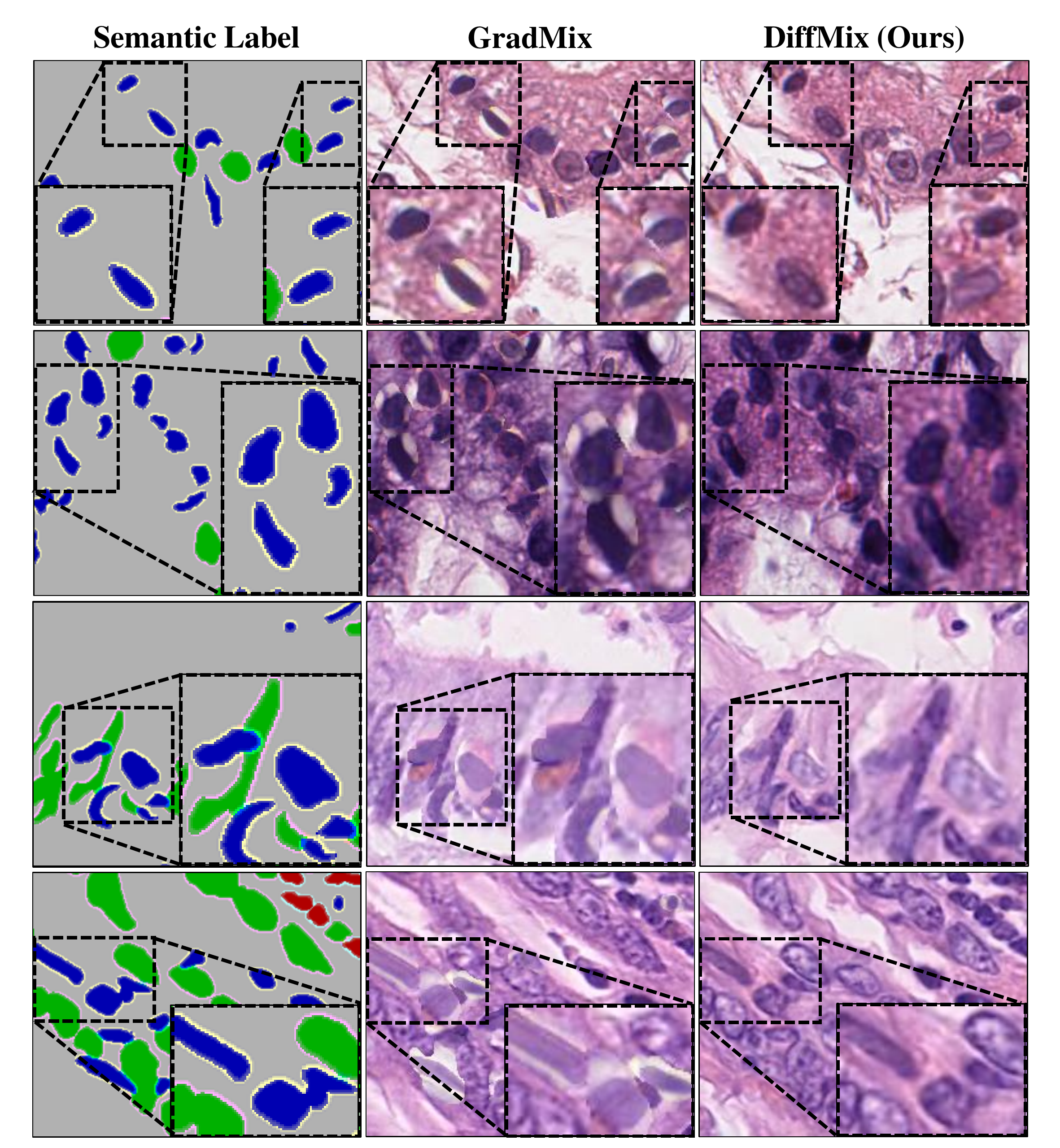}
    \caption{
        Synthesized patch examples are shown from left to right, displaying the semantic label, GradMix, and \texttt{DiffMix} (Ours). 
        The top two rows show the results of GLySAC, while the bottom two rows demonstrate the results of CoNSeP. %Note that \texttt{DiffMix} can generate more realistic synthetic images. 
        % Synthesized patch examples.
        % From left to right: Semantic label, GradMix, and \texttt{DiffMix} (Ours).
        % The top two rows show the results of GLySAC, and the bottom two rows display the results of CoNSeP.
    }
    \label{table:patch_examples}
\end{figure}

\end{document}